\documentclass[letterpaper, 10 pt, conference]{ieeeconf}  

\IEEEoverridecommandlockouts                              

\usepackage{changes}
\usepackage[T1]{fontenc}
\usepackage{times}
\usepackage{soul}
\usepackage{url}
\usepackage[hidelinks]{hyperref}
\usepackage{graphicx}
\usepackage{amsmath}
\usepackage{algorithm}
\usepackage{algpseudocode}
\usepackage{amssymb}
\usepackage{booktabs}
\usepackage{multirow}
\usepackage{array}
\usepackage{xcolor}
\usepackage{cite}
\usepackage{float}
\definecolor{purple}{RGB}{153, 0, 255}

\usepackage{hhline}
\usepackage[caption=false]{subfig}
\usepackage{listings}
\usepackage{fancyvrb}
\usepackage{xcolor}
\definecolor{highlight}{HTML}{DAE8FC}
\definecolor{blue}{HTML}{4a86e8}
\definecolor{yellow}{HTML}{e69138}
\definecolor{pink}{HTML}{e06666}

\usepackage{makecell}
\usepackage{threeparttable}
\usepackage{caption}
\title{\LARGE \bf
MoMa-Pos: An Efficient Object-Kinematic-Aware Base Placement \\Optimization Framework for Mobile Manipulation
}

\author{Beichen Shao$^{1\dagger}$, Nieqing Cao$^{2\dagger}$, Yan Ding$^{3*}$, Xingchen Wang$^{1}$, Fuqiang Gu$^{1}$,  Chao Chen$^{1*}$
\thanks{$^\dagger$ Equal Contribution, * Corresponding author}
\thanks{$^1$~College of Computer Science, Chongqing University, $^2$~Xi'an Jiaotong-Liverpool University, $^3$~Shanghai Artificial Intelligence Laboratory}
}

\begin{document}

\maketitle
\thispagestyle{empty}
\pagestyle{empty}

\begin{abstract}
In this work, we present MoMa-Pos, a framework that optimizes base placement for mobile manipulators, focusing on navigation-manipulation tasks in environments with both rigid and articulated objects.
Base placement is particularly critical in such environments, where improper positioning can severely hinder task execution if the object’s kinematics are not adequately accounted for.
MoMa-Pos selectively reconstructs the environment by prioritizing task-relevant key objects, enhancing computational efficiency and ensuring that only essential kinematic details are processed. 
The framework leverages a graph-based neural network to predict object importance, allowing for focused modeling while minimizing unnecessary computations. 
Additionally, MoMa-Pos integrates inverse reachability maps with environmental kinematic properties to identify feasible base positions tailored to the specific robot model.
Extensive evaluations demonstrate that MoMa-Pos outperforms existing methods in both real and simulated environments, offering improved efficiency, precision, and adaptability across diverse settings and robot models.
Supplementary material can be found at \url{https://yding25.com/MoMa-Pos}
\end{abstract}

\begin{figure*}[t]
    \centering
    \includegraphics[width=1\linewidth]{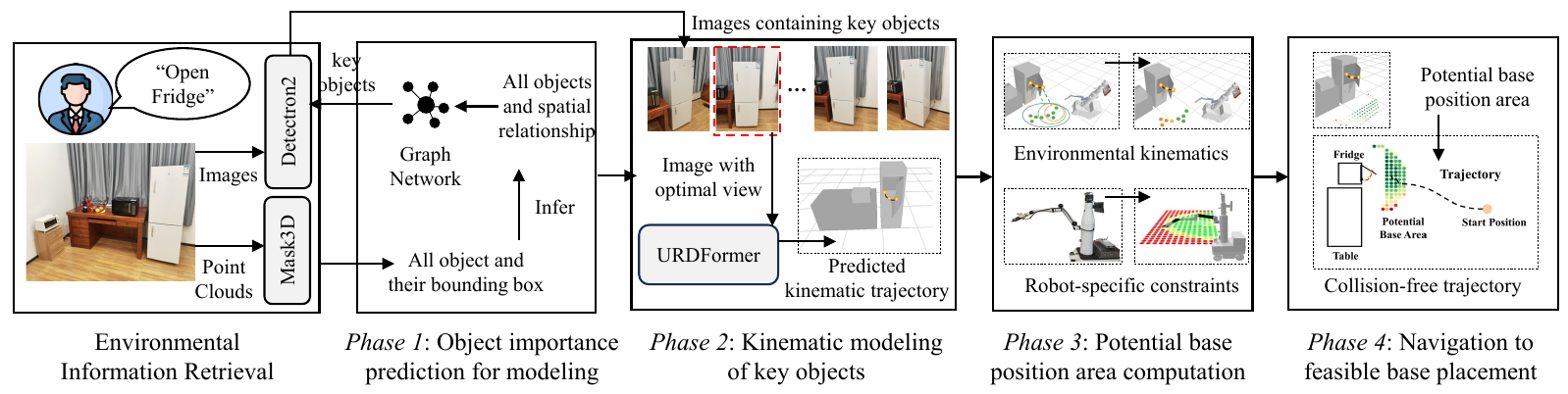}
    \caption{\textbf{Framework of MoMa-Pos}. The framework is composed of four key phases. 
    Phases 1 and 2 form the task-specific kinematic perceptual modeling, while Phases 3 and 4 focus on base placement optimization and guided navigation.
    In \emph{Phase 1}, key objects in the scene are prioritized to enable efficient kinematic modeling without processing every object.
    In \emph{Phase 2}, kinematic modeling is conducted for these prioritized objects, supporting the subsequent object-kinematic-aware base placement optimization.
    In \emph{Phase 3}, potential base placement areas are identified by considering both robot-specific constraints and environmental kinematics. 
    Finally, in \emph{Phase 4}, the robot navigates to the optimal position, ensuring physical feasibility and adaptability for task execution.
    \vspace{-1.em}
    }
    \label{fig:framework}
\end{figure*}

\section{Introduction}
Mobile manipulators, typically consisting of a mobile base and a robotic arm, are increasingly deployed in everyday environments like homes, where they are tasked with a wide range of navigation-manipulation tasks~\cite{yenamandra2023homerobot,habitatrearrangechallenge2022}.
These robots generally navigate to targets such as fridges and then execute manipulation tasks like opening fridge doors.
It is well-recognized that the success of these tasks is heavily dependent on the robots' base placement~\cite{gu2022multi,zhang2022visually,zhang2023base,sandakalum2022inv}.
Improper placement can significantly hinder the robots' ability to execute \emph{fixed-base manipulation}, where the mobile bases remain stationary while the robotic arms perform the manipulation tasks at hand.

However, determining feasible base placement remains challenging, particularly in household environments that contain both rigid and articulated objects. 
Physically interacting with articulated objects requires robots to be \emph{kinematically aware} of these objects' specific joint movements and constraints during manipulation.
For instance, when opening a fridge door, the robot must account for the door's rotational movement and position itself correctly to exert the necessary force along the appropriate kinematic trajectory.
Therefore, kinematic modeling of these articulated objects is essential.
While technologies in this area have advanced considerably~\cite{chen2024urdformer}, the associated computational costs remain significant.
The cumulative modeling time of all objects in a given scenario can be substantial, \emph{further diminishing the overall efficiency of robotic systems}.
Additionally, a feasible base placement is highly dependent on the \emph{robot model}.
Robots with shorter arms, such as Stretch~\cite{kemp2022design}, need to place themselves closer to target objects due to their limited arm reach.

Existing approaches in computing base position present several limitations.
The basic methods~\cite{szot2021habitat,puig2023habitat} set specific base placement areas empirically, without accounting for critical factors such as the kinematics of articulated objects and specific robot models, which limits their applicability in complex scenarios.
More advanced methods use inverse reachability maps~\cite{makhal2018reuleaux}, relying on precomputed kinematic solutions to guide base placement~\cite{vahrenkamp2013robot,zacharias2008positioning,makhal2018reuleaux,attamimi2012planning,zacharias2007capturing,reister2022combining}.
However, these approaches overlook the kinematic complexities of articulated objects, where complex motion and configuration dependencies impose constraints that extend beyond the limitations of reachability maps.
Moreover, these methods typically incur high computational costs, making them impractical for real-time applications.
Learning-based approaches predict feasible base placements by considering the kinematic properties of objects, the environment, and specific robot models~\cite{zhang2022visually,gu2022multi,gupta2023predictingmotionplansarticulating,zhang2023base,stulp2009action,honerkamp2021learning}.
However, they require extensive data collection and training, often tailored to predefined robot-environment pairs, which limits their generalizability to real-world applications.
Given these limitations, there is a pressing need for an efficient and generalizable framework that incorporates object-kinematic awareness and adaptability to various robot models.

This work introduces \emph{MoMa-Pos}, a framework designed to determine optimal base \underline{pos}itioning for \underline{mo}bile \underline{ma}nipulators prior to executing navigation-manipulation tasks.
A central aspect of MoMa-Pos is its ability to prioritize key objects in the scene, enabling efficient modeling without processing every object in the environment. 
By leveraging a graph-based neural network, our method predicts the importance of task-relevant objects, allowing for focused computation on critical elements. 
This selective modeling integrates advanced algorithms like URDFormer~\cite{chen2024urdformer}, facilitating accurate kinematic representations where necessary. 
The optimization of base positions in MoMa-Pos is distinguished by its integration of two critical factors: the use of inverse reachability maps to account for robot-specific constraints, and the incorporation of objects' kinematic properties from the environment to identify feasible base placements. 
This dual strategy refines potential base position areas, balancing physical feasibility with task-specific efficiency, leading to enhanced precision and adaptability.
Extensive experiments in both real and simulated environments demonstrate the superior performance of MoMa-Pos, consistently outperforming existing methods in terms of efficiency and accuracy across various scenarios, parameters, and robot models.

\begin{figure}[t]
    \centering
    \includegraphics[width=0.7\linewidth]{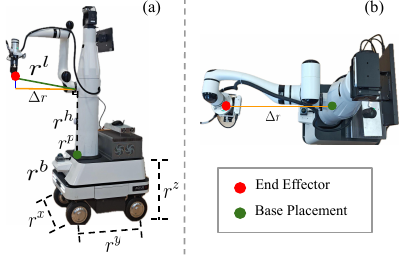}
    \vspace{-0.3em}
    \caption{Our mobile manipulator includes: a base, a body, and an arm. 
    The symbol $\Delta r$ represents the horizontal distance between the base's placement and the end-effector in the x-y plane. 
    $r^b$ represents the base dimensions, $r^h$ is the body height, and $r^l$ is the arm extension length.
    For explanations of additional symbols, see Section~\ref{sec:problem_setup}.}
    \label{fig:robot_model}
    \vspace{-1em}
\end{figure}

\section{Problem Statement}\label{sec:problem_setup}
In this section, we define the problem of computing the optimal base placement in an environment, denoted as $Env^O$, which contains a set of objects represented as $O=\{o_1, o_2, \cdots, o_N\}$.
Each object $o_i$ is characterized by its 3D position $o_i^p$ and 3D bounding box $o_i^{b}$.
The target object within these is specified as $o_t$, and the mobile manipulator is tasked with manipulating object $o_t$. 
This robot model, consisting of a base, a body, and an arm, is denoted by the tuple $(r^b, r^h, r^l)$, where $r^b$ represents the base dimensions $(r^x, r^y, r^z)$ (width, length, height), $r^h$ is the body height, and $r^l$ is the arm extension length.
The robot's 2D base position is denoted by $r^{p}$, which is assumed to be vertically aligned with the arm base.
The \emph{objective} is to determine a feasible base placement $r^{p*}$ for navigation-manipulation tasks.
The position must meet two critical criteria: the arm must be able to compute collision-free trajectories for task execution, and the base must avoid collisions with any objects in $Env^O$.

To obtain the position $o_i^p$ and bounding box $o_i^{b}$ for each object $o_i$, the robot first scans the environment to collect 3D point cloud data and videos.
The point cloud is processed by a pre-trained Mask3D model~\cite{Schult23ICRA} to segment objects and extract their corresponding 3D bounding boxes.
After removing outliers based on a distance threshold, the remaining points are used to compute an axis-aligned 3D bounding box, with $o_i^p$ defined as its geometric center. 
Additionally, the video data is used to select RGB images with optimal views for further kinematic modeling.

\section{The MoMa-Pos Framework}
The proposed MoMa-Pos consists of four phases, including object importance prediction for kinematic modeling, kinematic modeling of key objects, base placement optimization, and navigation to feasible base placement,
which are indicated in Fig.~\ref{fig:framework}.

\subsection{Object Importance Prediction for Kinematic Modeling}\label{sec:Predicting Object Importance for Environment Building}
The objective of object importance prediction is to select a subset of objects, $S$, from the object set $O$ for kinematic modeling, enabling efficient computation without processing every object. 
Rather than directly select the objects in an environment, we evaluate each object's importance and select those with an importance score exceeding a predefined threshold $\alpha$. 
This approach allows for flexibility; if the initial selection is insufficient, reducing $\alpha$ can include additional objects~\cite{silver2021planning}. 
We develop \emph{a prediction model}, $f: O \times O \rightarrow (0,1]$, where the output, $f(o_i, o_t)$, represents the likelihood of including object $o_i$ in $S$ based on its importance relative to the target object $o_t$.
Due to the challenges of obtaining precise labels for all objects in diverse environments\cite{silver2021planning}, we adopt an unsupervised learning approach, guided by the observation that an object's importance is often correlated with its spatial proximity to $o_t$ and its size, $o_i^{b}$. 
This correlation enables us to infer importance without exact labels. 

Our method structures within a graph-based architecture to calculate spatial proximity between objects $o_i$ and $o_t$ while considering object size $o_i^{b}$.
In $Env^O$, each object $o_i$ is represented as a node $v_i \in V$, with attributes such as size $o_i^{b}$. The spatial relationships between objects are captured by edges $E \subseteq V \times V$. 
An edge $(v_i, v_j)$ exists if object $o_i$ is spatially related to $o_j$ (e.g., on, in, or inside). 
The direction of the edge indicates spatial hierarchy, and the weight is defined as ${1}/{\textit{dist}_{xy}(o_i, o_j)}$, where $\textit{dist}_{xy}$ is the horizontal distance between $o_i$ and $o_j$.
To quantify the spatial proximity between nodes $v_i$ and $v_t$, we explored multiple algorithms, such as word2vec\cite{mikolov2013efficientestimationwordrepresentations}, which were found inadequate in capturing the spatial information.
After a thorough evaluation, we selected the DeepWalk algorithm~\cite{perozzi2014deepwalk} here.
Node sequences are generated through biased random walks, with the transition probability $P(v_i, v_j)$ weighted by the function $w(v_i, v_j) = k_0/\textit{dist}_{xy}(o_i, o_j) + (1-k_0) \times size(o_j)$, where $k_0$ is a tuning parameter. 
This bias ensures the walks reflect both spatial distance and object size, generating sequences that better capture object importance.
\begin{equation*}
P(v_i, v_j) = 
\begin{cases}
\frac{w(v_i, v_j)}{\sum_{u \in N(v_i)} w(v_i, u)}, & (v_i, v_j) \in E
\\ 0, & otherwise
\end{cases}\label{eqn:biased}
\end{equation*}
where $N(v_i)$ represents the set of nodes adjacent to $v_i$. 
After obtaining node embeddings, we compute the importance score of each object $o_i$ by calculating the cosine similarity~\cite{vijaymeena2016survey} between its embedding and that of the target object $v_t$. Objects with a score exceeding a threshold $\alpha$, where $\alpha \in [0.0, 1.0]$, are included in the set $S$. The threshold $\alpha$ is adjustable, starting at a higher value and decreasing if the system requires more objects to achieve optimal performance.

\subsection{Kinematic Modeling of Key Objects}\label{sec:Targeted Object Kinematic Modeling}

The purpose of this subsection is to conduct kinematic modeling of key objects within simulation environments.
These models will be utilized in the base placement computation for navigation planning.
The models will also facilitate the prediction of manipulation waypoints~\cite{xia2024kinematic} for articulated objects, enabling effective execution of real-world manipulation tasks.
To achieve this, we employ the state-of-the-art model, URDFormer, an end-to-end model based on Vision Transformers (ViT), which excels in capturing an object's kinematic structure through URDF models transformed from a single input image. 
However, \emph{URDFormer cannot directly ground the model to real-world objects}, often leading to distortions in size and pose when rescaling the detected objects. 

One of the reasons, discovered through our exploration, is that the input image’s viewpoint critically affects the modeling accuracy, with frontal views producing the best results. 
To address it, we design an algorithm to select the optimal viewpoint image of the target object from video streams, ensuring the input to URDFormer is optimal for more accurate kinematic modeling.
This process involves using Detectron2~\cite{wu2019detectron2} to detect the target object in each video frame, followed by sorting the frames based on the object’s central position in the image. 
We then fine-tuned a ResNet50 model~\cite{He16CVPR}, trained on a dataset we created, which includes 1,000 labeled images (frontal and non-frontal) of over 20 everyday objects such as refrigerators and microwaves, to classify whether the object in each frame is facing forward. 
The \emph{first best image}, which is classified as ``frontal,'' is selected as the input for URDFormer to generate an initial URDF model with interactive structures.

Furthermore, to address the inherent limitations of 2D single-image input which can often lead the generated URDF model to misalign the simulation with the real-world 3D environment, we correct the model using precise 3D information from point clouds. 
Specifically, we use 3D bounding boxes generated by Mask3D~\cite{Schult23ICRA} to align the object's size and pose, ensuring the URDF model accurately simulates an interactive environment that mirrors the real world. 
Finally, after inputting a recorded environment video, our algorithm selects the best frontal image of the target object and uses 3D bounding boxes to align the object's size and pose, generating an accurate URDF model. This allows us to precisely locate the actionable parts on articulated objects at any opening angle, ensuring alignment with real-world conditions. The continuous location information at each angle forms the object's manipulation waypoints.
Note that rigid objects do not require modeling with URDFormer. 
Instead, they can be approximated as bounding boxes, where the box dimensions correspond to the object's bounding size. 
This strategy significantly improves efficiency.

\begin{figure}[t]
    \centering
    \includegraphics[width=1.0\linewidth]{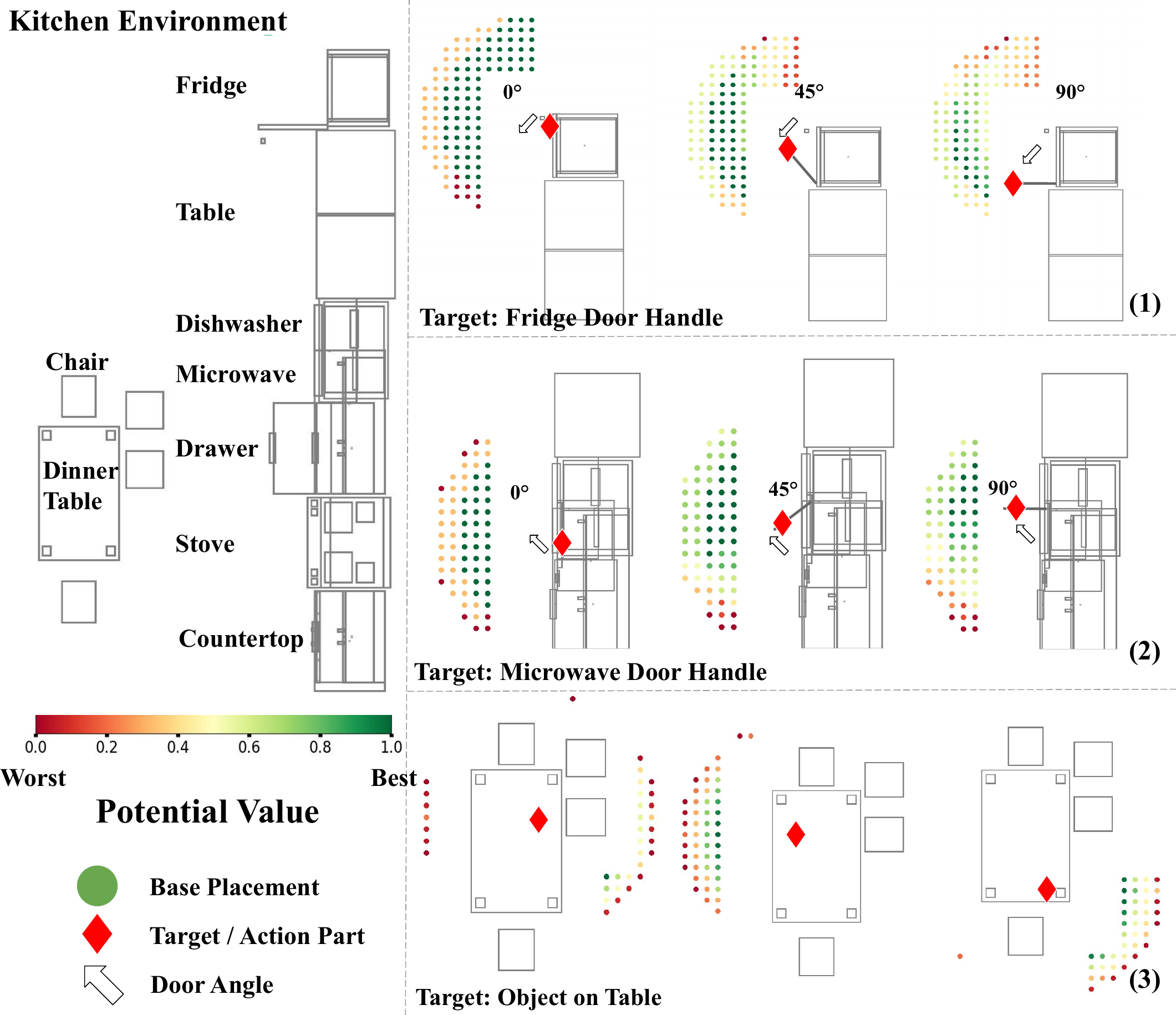}
    \caption{The upper-left figure provides an overview of the robots' operating environment, while the three subfigures on the right focus on key task-relevant objects in simulation. 
    Green circles indicate potential robot standing positions and red diamonds represent targets, such as the fridge door handle. 
    In example (1), when the door handle changes position, the corresponding feasible placement map is also altered, consistent with intuitive expectations. 
    This is best viewed in the enlarged digital version.}
    \label{fig:standingArea}
    \vspace{-1em}
\end{figure}

\subsection{Base Placement Optimization}\label{sec:compute_area}
To optimize the base placement, we first calculate the potential area for base positions (denoted as 
$A$), where there are no collisions with surrounding objects. We then refine this area by considering both the robot's constraints as well as the objects' kinematics in determining base placements.

To get an initial robot's potential base position area $A$, we consider both the robot's model $r^p$ and the target object's position $o_t^p$ to avoid any collisions.  
The potential area $A$ is calculated according to the following equation:
\begin{equation*}
\begin{split}
A = \left\{ r^p \mid \textit{dist}_{xy}(r^p,o_t^p) \leq \Delta r, \, \forall i, \, \neg overlap \left( r^p, r^b, o_i^p, o_i^b \right) \right\}
\end{split}
\end{equation*}
Here, $\Delta r$ represents the allowable horizontal distance between a base position $r^p$ and the target $o_t^p$, illustrated in an orange line in Fig.~\ref{fig:robot_model}.
The \text{overlap} function checks for collisions between the robot's base and surrounding objects $o_i$ by comparing their positions and boundaries.
More specifically, $\Delta r$ is calculated as $\sqrt{(r^l)^2-(o_t^{p.z} - r^{z} - r^{h})^2}$, where $r^l$ is the arm extension range (shown in green in the figure), derived from the robot's DH matrix.
The z-axis distance between $r^p$ and $o_t^{p.z}$ is indicated by $|o_t^{p.z} - r^{z} - r^{h}|$, marked as a blue line in the figure.

To further enhance the base placement optimization process, we integrate an inverse reachability matrix to account for robot-specific constraints and combine the potential field method to avoid potential interference with furniture and obstacles due to their possible kinematic trajectories. 
Utilizing the inverse reachability matrix, we can determine which points in $A$ a robot can reach given its joint configurations.
Calculated offline, the matrix allows us to precompute feasible base positions within area $A$.
By selecting base positions directly from the matrix rather than through random sampling, we ensure that each position satisfies the robot's kinematic constraints.
Inspired by the potential field method~\cite{ge2002dynamic}, we then We then assess the feasibility of each candidate position.  
The target object $o_t$ exerts an attractive force. 
For each candidate base position $r^p$, we sample a 3D position $(r^p, z)$ and connect it to $o_t^p$ with a line. 
If this line collides with any furniture’s bounding box, the potential value from the potential field $F(r^p, z)$ is set to zero, indicating the robot cannot reach the target from that position. 
Otherwise, we calculate $F(r^p, z) = 1 / \textit{dist}_{xy}(r^p, o_t^p)$. 
The final potential value is the sum of the value from the inverse reachability matrix and the value from the potential field.
Positions with higher final potential values are more feasible.
Additionally, we incorporate trajectory planning by discretizing it into multiple waypoints and calculating the potential for each, resulting in a more comprehensive potential field. 
The sum of the potentials along the trajectory produces the final potential map, which guides the selection of the optimal base placement. Fig.~\ref{fig:standingArea} illustrates the potential values for sampled positions in a kitchen environment.

\subsection{Navigation to Feasible Base Placement}\label{sec:identify_position}
Within area $A$, we use the Latin Hypercube sampling\cite{mckay1979lhs} method to sample $M$ 2D positions $r^p_i$, with higher potential values suggesting better feasibility. 
Instead of naively selecting the position with the lowest potential and checking feasibility via a motion planner like RRT*\cite{karaman2011sampling}, we aim to balance potential values with navigation costs. 
This leads to a problem resembling the Open Traveling Salesman Problem (TSP)\cite{chieng2014performance}, where the robot does not return to its starting position.

Given the large number of sampled positions, we group them into smaller sets, each containing $T$ candidates. 
Within each set, the optimal path is determined by solving the Open TSP, using edge weights that combine both distance $\textit{dist}_{xy}(r^p_i, r^p_j)$ and potential values. 
The edge weight is computed as $k_1 \times \textit{dist}(r^p_i, r^p_j) + k_1' \times \left( F(r^p_j) - F(r^p_i) \right)$, where $k_1$ and $k_1'$ are user-defined coefficients. 
Algorithms ranging from exact methods~\cite{boyd2007branch} to heuristic approaches~\cite{kizilatecs2013nearest} can be used, depending on the size of $T$. 
The search continues until a collision-free trajectory is found, and the corresponding base position $r^{p*}$ is selected.

\textbf{Lemma 1} (MoMa-Pos is complete). \emph{Given any object importance scorer $f: O \times O \rightarrow (0,1]$, threshold $\alpha\in [0.0, 1.0]$, and $M$ sampled positions, $M\in [1, \infty)$, if the motion planner PLAN is complete, then MoMa-Pos is complete.}

\begin{proof}
Given the codomain of $f$ excludes 0, an $\alpha > 0$ exists such that $\forall o \in O, f(o, o) \geq \alpha$.
With $M \rightarrow \infty$, a feasible position, if existent, will be sampled eventually.
Given PLAN's completeness, MoMa-Pos also achieves completeness. 
Hence, for any task and an infinite number of samples, MoMa-Pos will identify at least one solution, assuming one exists, through PLAN.
\end{proof}

\section{Experiments}

\begin{figure*}[t]
    \centering
    \includegraphics[width=1\linewidth]{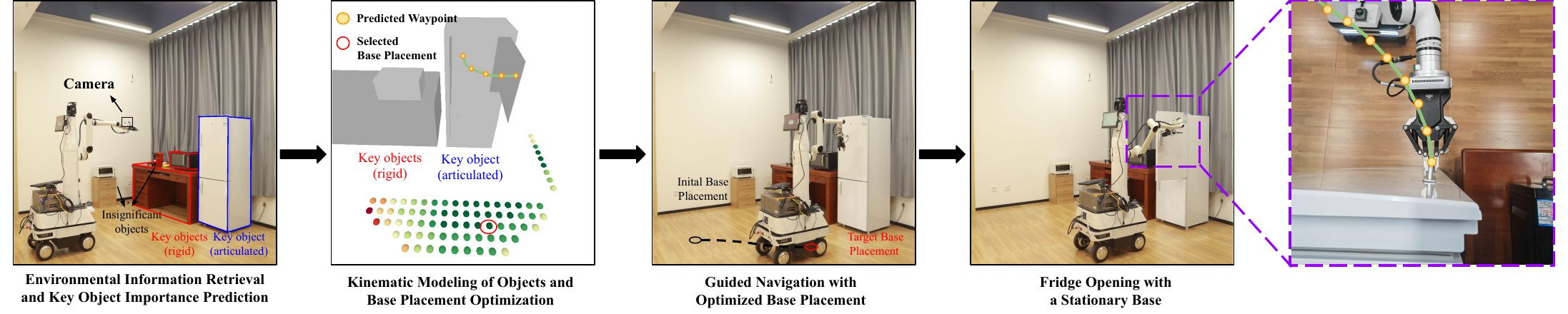}
    \caption{\textbf{Real Robot Demonstration.} We implemented MoMa-Pos on a real robotic system comprising an AgileX mobile platform and a Realman-63F robotic arm. 
    The robot's task in this setup is to open a fridge door. 
    In the first figure, MoMa-Pos predicts key objects in the environment, with these objects highlighted by colored outlines. 
    Meanwhile, the manipulation waypoints are predicted and indicated by yellow circles.
    Next, MoMa-Pos performs kinematic modeling, representing rigid objects as boxes and articulated objects using URDFormer. 
    A feasible base placement is then selected, taking into account robot-specific constraints and the kinematic properties of the environment. 
    The final two figures illustrate the robot successfully completing navigation, where the trajectory is indicated in a dashed black line, and fixed-base manipulation tasks using the predicted waypoints.}\label{fig:demo}
    \vspace{-1em}
\end{figure*}

In our experiments, we aim to address the following questions: 
\textbf{Q1:} How does MoMa-Pos's capability in determining feasible base placements? 
\textbf{Q2:} How important is the object importance predictor for efficiency in kinematic modeling? 
\textbf{Q3:} How accurate is the kinematic modeling?

\vspace{.5em}
\noindent
\textbf{Experimental Setup:}
In the \textbf{real robot} experiment, we construct a simplified home environment in a controlled laboratory setting, as illustrated in Fig.~\ref{fig:demo}.  
The scene includes five rigid objects (e.g., table, apple, and cup) and two articulated objects (i.e., fridge and microwave), with the rigid objects and microwave placed randomly on the table, which has a variable position, while the fridge remained fixed. 
The mobile manipulator, shown in Fig.~\ref{fig:robot_model}, consisted of an AgileX-based mobile platform and a Realman 63-F robotic arm, equipped with a Realsense 435 camera for RGB images, depth images, and point cloud data acquisition.
The mobile manipulator is tasked with grasping rigid objects (i.e., apple and cup) avoiding obstacles and opening articulated items.
For the fridge and microwave, we open them to 90 degrees during the experiments.
The methods for determining base placements are evaluated across five trials per task, with each scene featuring distinct object configurations.
The robot's initial position is kept consistent across all trials.
For manipulation, the spatial positions of the objects and their actionable parts (e.g., door handles) of articulated objects are determined using Mask3D~\cite{Schult23ICRA} and Detectron2~\cite{wu2019detectron2}. 
We then employ the RRT* algorithm, implemented via the open-source OMPL library~\cite{sucan2012open}, to compute corresponding manipulation trajectories.
For navigation, we utilize the Dynamic Window Approach (DWA)~\cite{fox1997dynamic}, chosen for its responsiveness and stability in complex dynamic environments, ensuring high precision and safety. 
A UWB-based solution~\cite{gezici2005localization}, relying on communication for global positioning, is employed, with an accuracy error of approximately 2 cm.

We also perform  \textbf{simulated robot experiments}. 
Unlike the real-world setup, the simulation involves a larger number of everyday objects (30, including a fridge, microwave, and cup) and significantly more tasks, totaling 500. 
In the simulation, the 3D positions of objects and their actionable components are directly accessible. 
The simulation tool is supported by the GitHub project\footnote{\url{https://github.com/AutonoBot-Lab/BestMan_Pybullet}}.
Our experiments were conducted on a computer with an AMD EPYC 7452 processor (32 cores, 64 threads), an Nvidia GTX 2080 GPU, and 128GB of RAM.
All code is publicly available for reproducibility at: ~\url{https://yding25.com/MoMa-Pos}

\begin{table*}[t]
    \scriptsize 
    \centering
    \renewcommand{\arraystretch}{0.7} 
    \fontsize{6}{10}\selectfont
    \begin{tabular}{
        l
        p{0.95cm}
        p{0.9cm}
        p{0.9cm}
        p{0.95cm}
        p{0.9cm}
        p{0.9cm}
        p{0.95cm}
        p{0.9cm}
        p{0.9cm}
        p{0.95cm}
        p{0.9cm}
        p{0.9cm}
    }
        
        \toprule
        {Methods} & \multicolumn{3}{c}{Grasp Apple \emph{(rigid)}} & \multicolumn{3}{c}{Grasp Cup \emph{(rigid)}} & \multicolumn{3}{c}{Open Fridge \emph{(right-hinged)}} & \multicolumn{3}{c}{Open Microwave \emph{(left-hinged)}} \\
        \cmidrule(lr){2-4} \cmidrule(lr){5-7} \cmidrule(lr){8-10} \cmidrule(lr){11-13}
        & {Time (s)} & {Cost (m)} & {SRate (\%)} & {Time (s)} & {Cost (m)} & {SRate (\%)} & {Time (s)} & {Cost (m)} & {SRate (\%)} & {Time (s)} & {Cost (m)} & {SRate (\%)} \\
        \midrule
        \emph{MoMa-Pos} (ours) & \textbf{13.5 $\pm$ 2.0} & {3.2 $\pm$ 1.2} & \textbf{100.0} & \textbf{13.2 $\pm$ 2.4} & {3.5 $\pm$ 1.4} & \textbf{100.0} & {17.4 $\pm$ 2.3} & {2.3 $\pm$ 0.1} & \textbf{100.0} & {17.2 $\pm$ 1.0} & {3.7 $\pm$ 1.6} & \textbf{100.0} \\
        \emph{Habitat} & {17.3 $\pm$ 0.2} & \textbf{3.0 $\pm$ 0.0} & {100.0} & {16.9 $\pm$ 2.7} & \textbf{3.2 $\pm$ 0.0} & {100.0} & \textbf{15.7 $\pm$ 1.4} & \textbf{2.2 $\pm$ 0.0} & {18.0} & \textbf{14.4 $\pm$ 0.4} & \textbf{3.2 $\pm$ 0.0} & {22.0} \\
        \emph{M3*}              & {17.8 $\pm$ 2.3} & {3.2 $\pm$ 1.1} & {100.0} & {17.3 $\pm$ 3.2} & {3.3 $\pm$ 0.1} & {100.0} & {18.2 $\pm$ 8.4} & {2.2 $\pm$ 0.1} & {26.0} & {17.4 $\pm$ 1.4} & {3.5 $\pm$ 0.0} & {44.0} \\
        \emph{Reuleaux}        & {19.2 $\pm$ 4.3} & {3.6 $\pm$ 1.6} & {100.0} & {18.7 $\pm$ 3.9} & {3.7 $\pm$ 1.1} & {100.0} & {17.7 $\pm$ 2.7} & {2.5 $\pm$ 0.1} & {70.0} & {18.0 $\pm$ 2.0} & {3.6 $\pm$ 1.4} & {82.0} \\
    \bottomrule
    \end{tabular}
    \caption{A comparison between MoMa-Pos and baseline methods in the real robot experiments in terms of task execution time (Time), navigation cost (Cost), and success rate (SRate) across different furniture categories, namely containers and non-containers. 
    The optimal results are highlighted in bold.
    }
    \label{tab:main}
    \vspace{-1em}
\end{table*}


\vspace{.5em}
\noindent
\textbf{Baselines:} We compare our method with the following reimplemented baselines, which are inspired by original data-driven approaches:
\begin{itemize}
    \item \textbf{Habitat}~\cite{habitatrearrangechallenge2022}: For rigid objects such as tables, the base placement closest to the target is selected, ensuring it is navigable. 
    For articulated objects like fridges, the base placement is determined based on a fixed position relative to these objects, guided by prior experience, while also ensuring no collisions with the environment.
    \item \textbf{M3*}~\cite{gu2022multi}: Based on the original method, a designated area is divided into $5\times5$ cm cells. 
    For rigid objects, the closest navigable position to the target is selected, prioritizing navigation cost. For articulated objects, base positions are randomly sampled within the area until a feasible placement is found. 
    Unlike the original, which employs reinforcement learning to optimize the initial base placement, this baseline lacks such optimization, potentially leading to higher navigation costs.
    We use * to denote differences.
    Additionally, while the original method allows for dynamic base repositioning during manipulation to enhance grasping, our implementation uses fixed-base manipulation, which introduces limitations and requires adjustments to accommodate our experimental setup.
    \item \textbf{Reuleaux}~\cite{makhal2018reuleaux}: Similar to M3*, it starts by designating a specific area, divided into $5\times5$ cm cells.
    It then evaluates how easily the robot can reach the target without obstacles. 
    The robot chooses its base placement randomly, prioritizing those with the highest reachability to ensure optimal accessibility.
\end{itemize}
Different from MoMa-Pos, none of these three baselines account for object kinematics, which can lead to suboptimal manipulation performance, particularly when handling articulated objects.

\vspace{.5em}
\noindent
\textbf{Rating Criteria:} 
To ensure a fair comparison, all baselines are evaluated within a complete working environment. 
These methods do not account for the kinematic information of articulated objects, treating all objects as simple bounding boxes, thus saving time on kinematic modeling.
We evaluate the performance of all approaches using the following metrics: \emph{task execution time} ({Time}), measured in seconds; \emph{navigation cost} ({Cost}), measured in meters; and \emph{success rate} ({SRate}).
For a more detailed breakdown of time consumption in determining feasible positions, we measure the time spent in each key step of the process, which includes three stages: predicting object importance, performing kinematic modeling, and optimizing base placement. 
\emph{The time allocated for modeling} not only includes the modeling process itself but also the integration of modeled objects into the simulator.
To focus on the efficiency of the task's core elements, we exclude navigation time from task execution time, as robot movement speed can introduce significant variability unrelated to task completion efficiency. 
Instead, the navigation cost metric evaluates spatial movement efficiency, ensuring that it does not interfere with the direct assessment of task performance.

\subsection{Results and Analysis}

\subsubsection{MoMa-Pos vs. Baselines} 
We aim to address \textbf{Q1} here.
The main results are presented in TABLE~\ref{tab:main}.
The table demonstrates that our method consistently achieves a 100\% success rate across all trials in the working environment.
Compared to the baseline methods (Habitat, M3*, and Reuleaux), the success rate of MoMa-Pos is significantly higher, particularly with articulated objects such as fridges.
This is due to our method’s ability to account for the structure of the objects, effectively avoiding obstructions.
MoMa-Pos also excels in efficiency, outperforming the M3* and Reuleaux baselines.
These methods rely on random sampling within a predefined area and often fail to quickly locate a viable position, leading to increased time costs and inconsistent performance, as evidenced by the high variance in M3* and Reuleaux's results.
Although Habitat is sometimes more efficient, its lower success rate, especially with articulated objects, limits its practical application.
Our method not only demonstrates significant advantages in success rate and time efficiency with articulated objects but also ensures the low navigation cost for finding feasible solutions.
In our tests, MoMa-Pos consistently identifies a feasible solution within 18 seconds.



\subsubsection{Object Importance Predictor for Efficiency in Kinematic Modelling} 
\begin{figure}[h]
    \centering
    \includegraphics[width=0.8\linewidth]{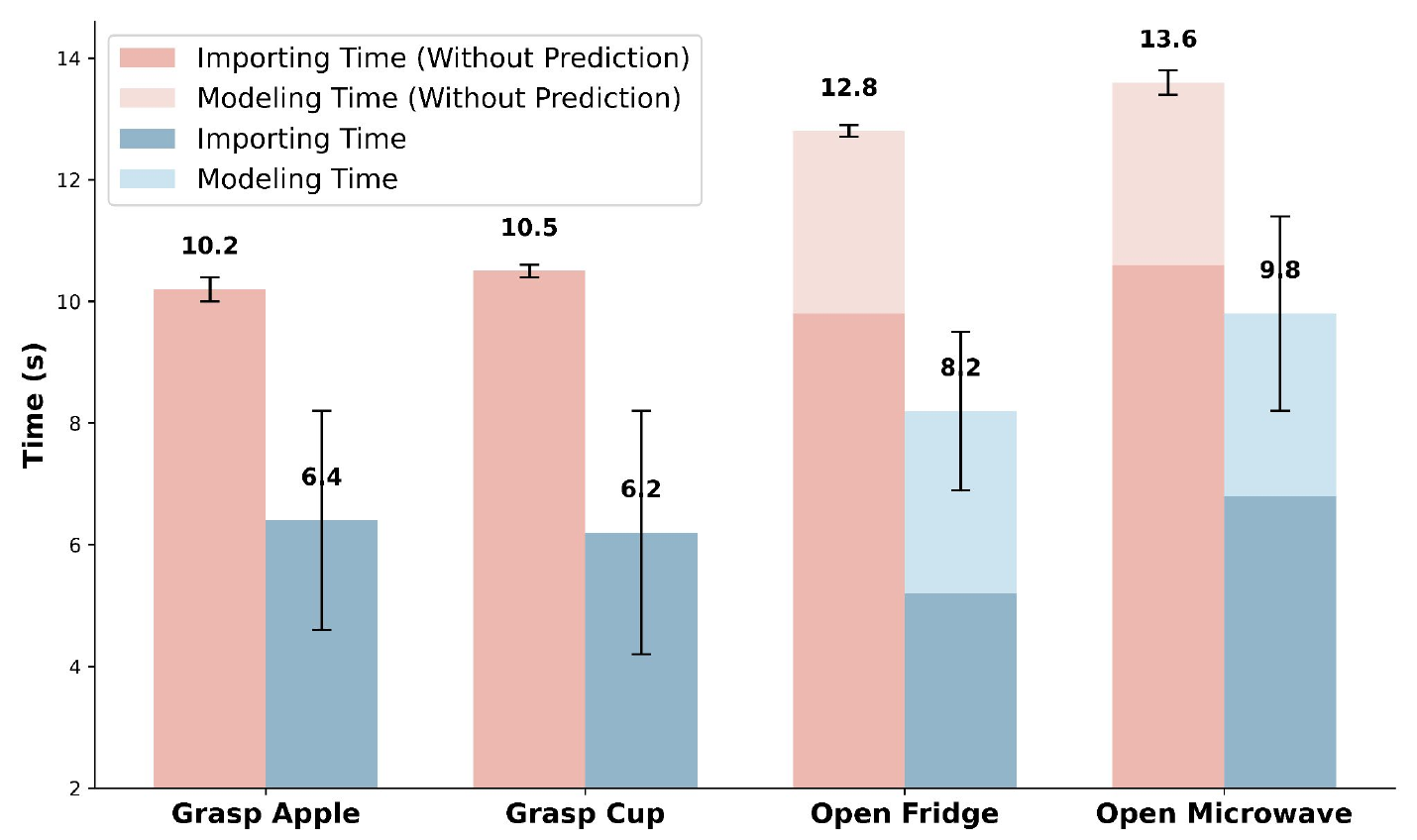}
    \caption{The modeling time comparison between MoMa-Pos and its variant,
which lacks the object importance prediction, with the x-axis
denoting the location of the target object.}
    \label{fig:prediction}
\end{figure}

\textbf{Q2} is answered here.
Our experimentation begins by setting the threshold value $\alpha$ (specifically, 0.45) within the object importance prediction module.
The achieved success rate of 100\% confirms that no crucial object has been overlooked. 
Fig.~\ref{fig:prediction} illustrates the modelling time for MoMa-Pos and a variant devoid of the prediction module.
It is clear that incorporating prediction in MoMa-Pos markedly decreases the time cost. 
Additionally, while the URDFormer-based modeling time for the cup is significantly shorter, the overall time is prolonged due to the increased time required for importing objects into simulators, as the cup's complex surroundings demand more processing compared to the fridge.

\subsubsection{Kinematic Modeling Accuracy} 
\textbf{Q3} is addressed here.
TABLE~\ref{tab:kinematic} presents the accuracy of kinematic modeling. 
For the Vanilla URDFormer, the generated object model exhibits significant misalignment with the real-world object, resulting in a spatial error of approximately 0.38m, as shown in the table. 
The spatial error is calculated based on AP and the fridge dimensions, with the unit in meters. 
To improve alignment with real-world objects, we apply two methods to obtain the bounding box: Omni3D~\cite{ brazil2023omni3dlargebenchmarkmodel} and Mask3D. 
These methods differ in their approach, with Omni3D utilizing RGB images and Mask3D leveraging depth data. 
Compared to Omni3D, Mask3D demonstrates significantly higher accuracy, reducing the modeling error to 0.04m.

\begin{table}[h]
    \centering
    \renewcommand{\arraystretch}{1}
    \begin{tabular}{
    l
    >{\centering\arraybackslash}p{4cm}}
    \toprule
    {Methods} & {Modeling Accuracy Error (m)} \\
    \midrule
    Vanilla URDFormer & 0.38 \\
    URDFormer\texttt{+}Omni3D & 0.13 \\
    URDFormer\texttt{+}Mask3D (ours) & 0.04 \\
    \bottomrule
    \end{tabular}
    \caption{Comparison of modeling accuracy errors across different methods for fridge modeling.}
    \label{tab:kinematic}
    \vspace{-0.5em}
\end{table}

We also calculate the average time distribution across all tasks, including those involving both rigid and articulated objects, throughout the entire process.
TABLE~\ref{tab:time_distribution} presents the results. 
It is clear that the majority of the time (over 98\% of the total time cost) is dedicated to kinematic modeling and determining feasibility through a sampling-based motion planner, such as RRT*. In contrast, the time spent on candidate selection, including solving the Open Traveling Salesman Problem (TSP), is minimal, accounting for less than 2\% of the total time (less than 0.3 seconds).


\begin{table}[h]
    \centering
    \scriptsize
    \begin{tabular}{lc}
        \toprule
        \textbf{Process} & \textbf{Time Distribution (\%)} \\
        \midrule
        Modeling (including importing time) & 58.0 \\
        Checking feasibility of samples (OMPL) & 40.0 \\
        Potential field computation & 0.9 \\
        Importance prediction & 0.8 \\
        \bottomrule
    \end{tabular}
    \caption{Average time distribution of entire process in MoMa-Pos}
    \label{tab:time_distribution}
\end{table}

Our framework is further evaluated using three simulated robots: Fetch, Segway+UR5e, and Husky+UR5e. 
In all trials, MoMa-Pos achieves a 100\% success rate, demonstrating its robustness across different robot models. 
This highlights the versatility of our method, which is adaptable to various scenarios. 
Compared to baselines, MoMa-Pos significantly reduces the effort needed for integrating kinematic models into the simulator. 
Note that the simulated experiments do not involve kinematic modeling steps.
Additional simulation results are provided in the appendix\footnote{\url{https://yding25.com/MoMa-Pos/Appex}}.

\section{Conclusion}
In this work, we introduce MoMa-Pos, a novel framework designed to optimize base placement for mobile manipulators, particularly in environments containing both rigid and articulated objects. 
MoMa-Pos addresses key challenges in navigation-manipulation tasks by selectively modeling task-relevant objects, leveraging a graph-based neural network to enhance computational efficiency. 
By integrating inverse reachability maps and environmental kinematic properties, the framework enables precise base placement tailored to specific robot models, ensuring adaptability and physical feasibility across diverse environments.

\bibliographystyle{IEEEtran}
\bibliography{references}

\end{document}